\documentclass{article}
\usepackage{spconf,amsmath,graphicx}
\usepackage[numbers,sort]{natbib}
\usepackage{graphics} 
\usepackage{float}
\usepackage{amsmath}
\usepackage{amssymb}
\title{BFRFormer: Transformer-based generator for Real-World Blind Face Restoration}

\name{
\parbox{\linewidth}{\centering
Guojing Ge\(^{1,2,*}\), Qi Song\(^{3,*}\), Guibo Zhu\(^{1,2,5,6,\dag}\), Yuting Zhang\(^4\), Jinglu Chen\(^1\), \\ Miao Xin\(^{1}\), Ming Tang\(^{1}\), and Jinqiao Wang\(^{1,2,6}\)
}
\thanks{\noindent* Equal contribution.  \dag \ Corresponding author. \\ \
This work was supported in part by the National Key RD Program of China (No.2022ZD0160601), National Natural Science Foundation of China (No.62276260,62076235,61906195), Beijing Municipal Science and Technology Project (Z231100007423004), sponsored by Zhejiang Lab (No.2021KH0AB07).}
}

\address{\(^1\)Institute of Automation, Chinese Academy of Sciences  \(^2\)Wuhan AI Research  \\
\(^3\)Hong Kong Baptist University 
\(^4\)China Telecom Corporation Ltd \\  \(^5\)Shanghai Artificial Intelligence Laboratory 
\(^6\)University of Chinese Academy of Sciences}

\begin{document}

\maketitle

\begin{abstract}
Blind face restoration is a challenging task due to the unknown and complex degradation. Although face prior-based methods and reference-based methods have recently demonstrated high-quality results, the restored images tend to contain over-smoothed results and lose identity-preserved details when the degradation is severe. It is observed that this is attributed to short-range dependencies, the intrinsic limitation of convolutional neural networks. To model long-range dependencies, we propose a Transformer-based blind face restoration method, named BFRFormer, to reconstruct images with more identity-preserved details in an end-to-end manner. In BFRFormer, to remove blocking artifacts, the wavelet discriminator and aggregated attention module are developed, and spectral normalization and balanced consistency regulation are adaptively applied to address the training instability and over-fitting problem, respectively. Extensive experiments show that our method outperforms state-of-the-art methods on a synthetic dataset and four real-world datasets. The source code, Casia-Test dataset, and pre-trained models is released at https://github.com/s8Znk/BFRFormer.
\end{abstract}

\section{Introduction}
Blind face restoration aims at recovering high-quality face images from low-quality counterparts that suffer from unknown degradation. There are many applications such as old photo restoration and low-quality face recognition. The authentic, real-world low-quality images usually contain complex and diverse distributions that are impractical to imitate. For example, in the old photo restoration tasks, spatial uniformity and color fading are the major difficulties.

The current mainstream approaches are geometric prior methods~\cite{ChenPSFRGAN}\cite{2017FSRNet}\cite{2019Progressive}\cite{2018Deep}\cite{2018Face}\cite{2018Super}\cite{2016}, reference-based methods~\cite{2019Exemplar}\cite{2020Blind}
\cite{DR2}\cite{2020Enhanced}\cite{2018Learning}\cite{2022RestoreFormer}\cite{gu2022vqfr}\cite{li2022learning} and GAN prior-based methods~\cite{2021gpen}
\cite{wang2021gfpgan}. Geometric priors can be facial landmarks~\cite{2017FSRNet}\cite{2019Progressive}, facial parsing maps~\cite{ChenPSFRGAN}\cite{2018Deep}, or facial component heatmaps~\cite{2018Face}. However, those priors are estimated from degraded images and cannot offer accurate inputs when the degradation is severe. Besides, the geometric structures cannot provide sufficient information to recover facial details. The limitation of reference-based methods is that the learned codebook usually contains general characteristics of the training data set, losing identity-preserved details. 
 \begin{figure*}[ht]
    \centering
     \includegraphics[width=0.9\textwidth]{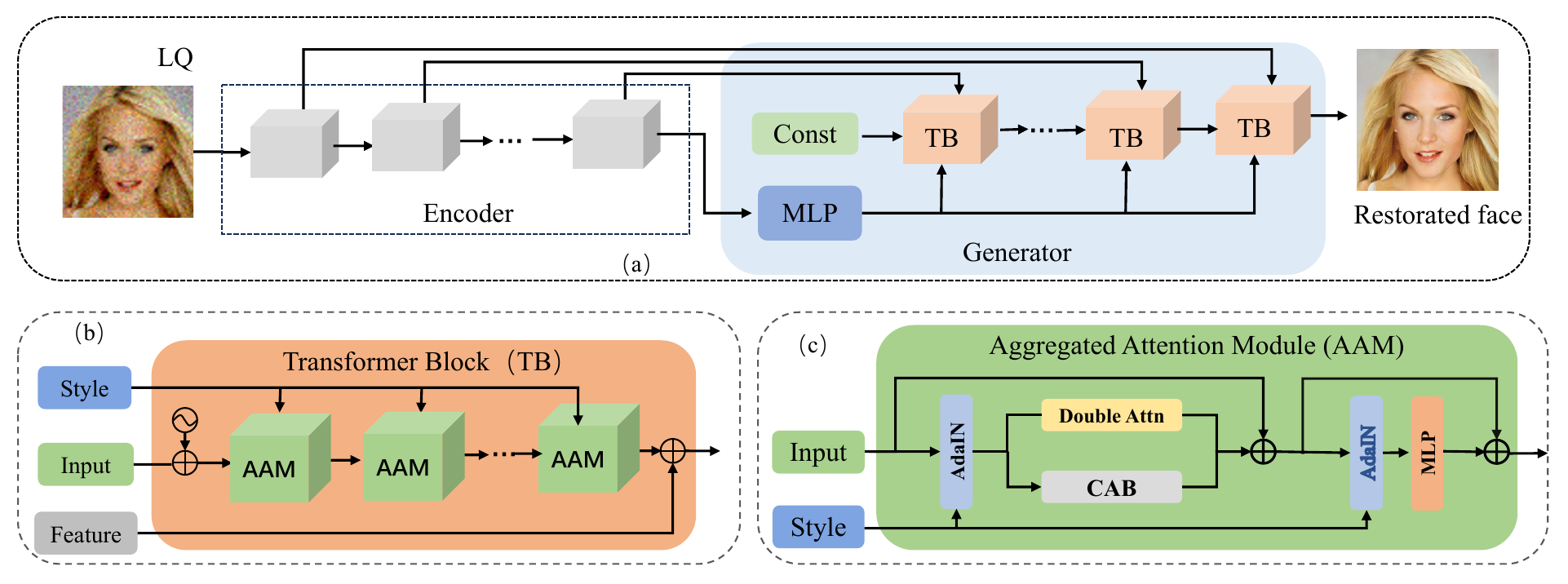}
    \vspace{-0.32cm}
    \caption{Overview of BFRFormer framework. (a) It follows an encoder and generator architecture; (b) TB is a transform-based block of the generator; (c) Aggregated Attention Module (AAM) which combing channel attention extracting global information with double-attention extracting local information, to activate more input pixels for face restoration.}
    \label{fig:framework}
\end{figure*}
GAN prior-based methods, instead of the previous facial priors such as geometric priors, have recently demonstrated high-quality results in this task. GFPGAN~\cite{wang2021gfpgan} and GPEN~\cite{2021gpen} extract local detail information and global identity information from low-quality input and then leverage the pre-trained GAN as a generator, achieving a good balance between visual quality and fidelity. Although existing methods have recently demonstrated high-quality results for blind face restoration, the restoration images tend to generate over-smoothed results and lose the identity-preserved details especially when the degradation is severe. 

To address the above problems, we develop a Transformer-based blind face restoration method, BFRFormer, to take advantage of both GAN prior-based methods and Transformer, yet without geometric prior or reference prior. Compared with other embedded CNN in the GAN prior-based methods, the Transformer-based generator can effectively model long-range dependency, improving the over-smoothed problem and the identity-preserved details. In BFRFormer, to remove blocking artifacts and generate realistic face details, the wavelet discriminator and an aggregated attention module are developed, respectively. The wavelet discriminator suppresses the artifacts by the periodic artifact pattern which can be easily distinguished in the spectral domain. The aggregated attention module (AAM), which contains channel attention and double attention, aims to enlarge the effective receptive field of the low-quality input. To address training instability and over-fitting problems, spectral normalization (SN) and balanced Consistency Regulation (bCR) are adaptively applied. Existing test datasets, such as the CelebChild-Test containing 180 child faces of celebrities and the WebPhoto-Test consisting of 407 real-life faces, are not diverse or fair enough to represent real-world low-quality images. In order to fairly evaluate the generalization of the existing blind face restoration methods, we construct a new real-world test dataset. Additionally, we experimentally show that perception loss, facial ROI discriminator, and data augmentation can improve performance well. 

In summary, this paper makes the following contributions: 

\textbf{•} We develop a Transformer-based training method embedded in the GAN prior-based framework for blind face restoration in an end-to-end manner compared with Coderformer and VQFR. 

\textbf{•} We propose a novel aggregated attention module, combining channel attention extracting global information and double-attention extracting local information, to activate more effective pixels of the low-quality input. 

\textbf{•} We construct one large test benchmark considering wider variation than the existing test datasets, including different ethnicities, different ages, different occlusion, and more than ten thousand persons for diversity. 

\textbf{•} Extensive experiments show that our method outperforms state-of-the-art methods on a synthetic dataset and four real-world datasets.

\section{Method} In this section, we will describe the overall pipeline, the details of the encoder, generator, and the loss function. 
\vspace{-10pt} 
\subsection{Overall Pipeline}
Given a low-quality face image, the purpose of blind face restoration is to reconstruct the high-quality image with realistic facial details while reserving identity information of the low-quality face image. The overall framework of BFRFormer is depicted in Figure \ref{fig:framework}. It comprises three modules: encoder, generator, and loss function. 
\vspace{-10pt}  
\subsection{Encoder network}
The encoder network is a simple convolution network, compared with RRDBNet~\cite{2018ESRGAN_RRDBNET} used in GLEN~\cite{2021GLEAN} and VQFR~\cite{gu2022vqfr}. The encoder extracts the multi-resolution features containing shallow features and deep features of the input image. %The shallow features only include the Blur layer, Conv layer, and FusedLeakyReLU layer. The deep feature captures the identity information of the images.
\vspace{-10pt}  
\subsection{Generator network}
As shown in Figure \ref{fig:framework}, the generator adapts a style-based architecture
~\cite{2019A}\cite{2021StyleSwin}\cite{qi2023degradation} that takes shallow features and style vectors as inputs where style vectors are injected into each Transformer Block (TB). The multi-level shallow features can add or concatenate to the output TB. Furthermore, the shallow features are concatenated rather than added to the Aggregated Attention Module (AAM) output. 
\vspace{-10pt} 
\subsubsection{Transformer Block (TB)}
The Transformer Block (TB) is the basic building block of the generator. The input features of TB are the output of the previous TB, shallow features achieved from the encoder, and style vector achieved from MLP. The input feature X is first processed by the AAM model with the style vector, and then the features obtained from the encoder are concatenated with the feature from AAM. Specifically, we have
\vspace{-0.1cm}
\begin{equation}
\begin{aligned}
     &X=\mathrm{AAM}(X,Style), 
\end{aligned}
\end{equation}
\vspace{-0.7cm}
\begin{equation}
\begin{aligned}
    &X=\operatorname{Concat}(Feature,X),  
\end{aligned}
\end{equation}
\vspace{-0.6cm}
\begin{equation}
\begin{aligned}
    &Y=\operatorname{Upsample}(X),
\end{aligned}
\end{equation} 
where the concatenated are further Up-sampled.
\vspace{-10pt} 
\subsubsection{Aggregated Attention Module (AAM)}
A channel attention block in parallel with the double-attention followed by AdaIN~\cite{2017adain}. To avoid the possible conflict of channel attention block and double attention on optimization and visual representation, a small constant \( \alpha \) is multiplied by the output of CAB. The whole process of Aggregated Attention Module (AAM) is computed as:
% \vspace{-0.2cm}
\begin{equation}
\begin{aligned}
    &X_{N}=\mathrm{AdaIN}(X_{N},Style),
\end{aligned}
\vspace{-0.25cm}
\end{equation}
\begin{equation}
\begin{aligned}
    &X_{M}=\mathrm{DoubleA}\left(X_{N}\right) +\alpha \mathrm{CAB}\left(X_{N}\right) + X,
\end{aligned}
\end{equation}  
\vspace{-0.55cm}
\begin{equation}
\begin{aligned}
    &X=\operatorname{MLP}(\operatorname{AdaIN}(X_{M},Style))+X_{M},    \\
\end{aligned}
\end{equation}

\noindent where $X_{N}$ and $X_{M}$ denote the intermediate features. Y represents the output of
AAM.  MLP denotes a multi-layer perception. 
\vspace{-10pt} 
\subsubsection{Channel Attention Block}
 A CAB consists of two standard convolution layers with a GELU activation function between them and a channel attention (CA) module. The details is shown in Figure \ref{fig:framework}.
\subsubsection{Double Attention}
Double attention~\cite{2021StyleSwin} aims to achieve an enlarged receptive field and allows a single Transformer block to simultaneously achieve the context of the local and shifted windows.
\vspace{-10pt} 
\subsection{Loss Functions}
In the Transformer-based GAN training process, the loss function is extremely important for the stability of training. Our losses involve several aspects, including pixel-level, component-level, and image-level.
In the training process of baseline, perceptual loss, and component-level loss are not used. The detailed discussions are as following.

\noindent \textbf{Pixel-level loss}.
L1 loss and perceptual loss are used in this paper. L2 loss is verified that it can not deal with high-frequency content in the image, resulting in overly smooth images~\cite{ChenPSFRGAN}. The perceptual loss further improves the image quality as shown in Table \ref{tab:ablation_study}. 

\noindent \textbf{Component-level loss}. Following~\cite{wang2021gfpgan}, we only focus on regions $r$ \( \epsilon \) $\{$ eyes, mouth $\}$. Specifically, the loss functions are formulated with discriminative loss 
\vspace{-0.1cm}
\begin{equation}
    \mathcal{L}_{roi} = \sum_{r\in ROI}E_{\hat{\boldsymbol{y}}_\mathrm{r}}\left[\log\left(1-D_\mathrm{r}\left(\hat{\boldsymbol{y}}_\mathrm{r}\right)\right)\right]. 
\end{equation}

\noindent \textbf{Image-level loss} Adversarial loss and identity preserving loss are used as image-level loss. We employ the wavelet discriminator in SWAGAN~\cite{gal2021swagan} to solve the blocking artifacts.

\begin{equation}
\mathcal{L}_{a d v}=-\mathbb{E}_{x}\left[\operatorname{softplus}\left(D_g\left(x\right)\right)\right]
\end{equation}

\section{Experiments}
\subsection{Datasets and Evaluation Metric}

To evaluate our model, we use one synthetic dataset~\cite{2017Progressive} and
four different real-world datasets to compare the proposed method with other blind face restoration methods.  CelebA-Test is the synthetic dataset with $5,000$ CelebA-HQ images. The generation way is the same as that during training.  \\
%\vspace{-10pt} 
\textbf{Casia-Test.} CASIA dataset~\cite{2014Learning} contains low-quality images in the wild, containing $10,575$ subjects and $494,414$ images. We randomly select three photos for each subject and use a quality assessment algorithm to compute the score of each face image. The lowest two images are selected as our target images. Among these, too many large poses are selected, and four experts are selected to balance the dataset to make it more impartial. At last, we get $16,683$ images containing $10,575$ identities to guarantee diversity with different races, ages, extreme poses, and light conditions.  

 \begin{figure*}
  \centering
    % \fbox{\rule{0pt}{2in} \rule{.9\linewidth}{0pt}}
    \includegraphics[width=2.8cm]{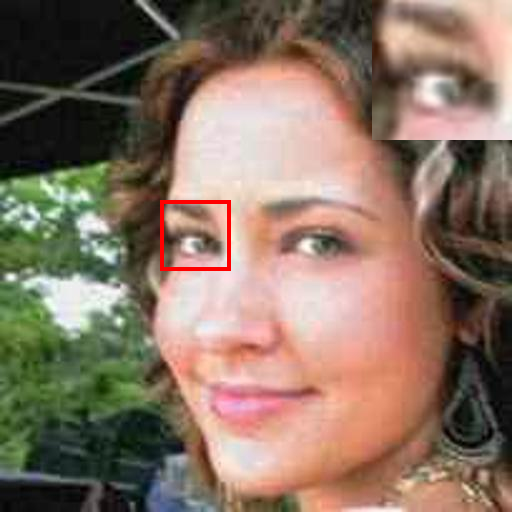}
    \includegraphics[width=2.8cm]{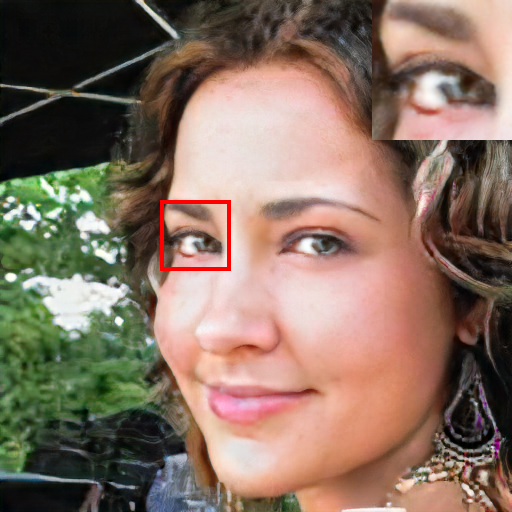}
    \includegraphics[width=2.8cm]{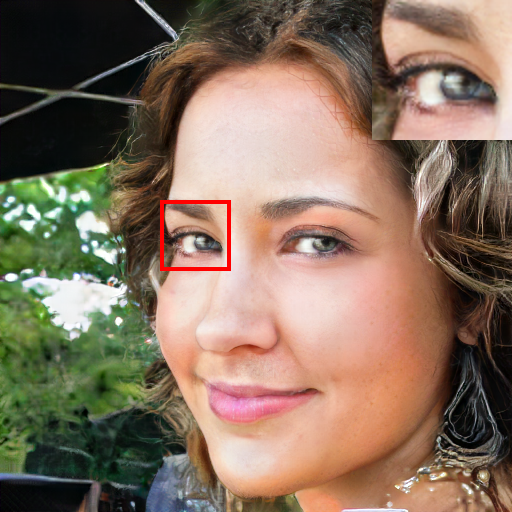}
    \includegraphics[width=2.8cm]{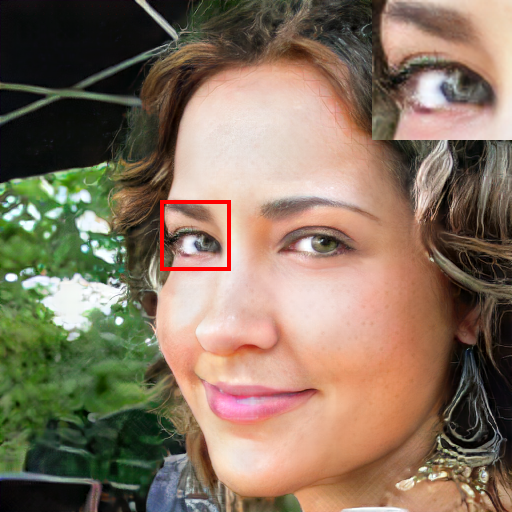}
    \includegraphics[width=2.8cm]{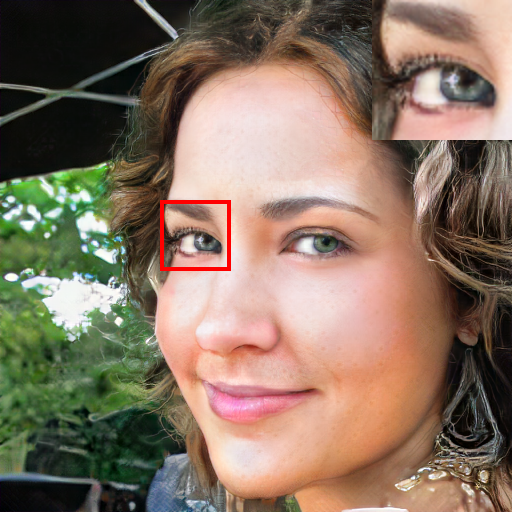}
    \includegraphics[width=2.8cm]{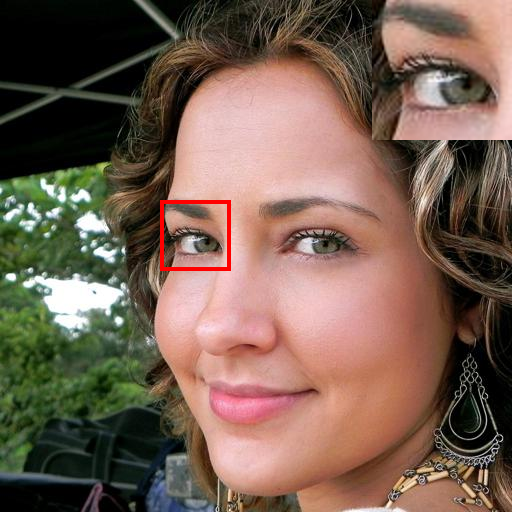}
    \vspace{-0.25cm}
    \caption{Comparison of our variants of BFRFormer. (a) Low-quality input; (b) Perceptual Loss;  (c) bCR; (d) CAB; (e) Facial Component; (f) Ground Truth. }
  \label{fig:abality}
\end{figure*}

    \vspace{-0.2cm}
    
\begin{table*}
    \centering
     %\label{tab:real-world-comparision}
    \setlength{\tabcolsep}{5mm}{}
    \begin{tabular}{l|l|l|l|l}
    \hline
        Datasets &  \ \textbf{ LFW-Test}   &  \ \  \textbf{CelebChild} &  \ \  \textbf{WebPhoto} &   \ \ \textbf{Casia-Test}  \\
        Method &  \ FID$\downarrow $\quad \text{NIQE}$\downarrow$	 & \ FID $\downarrow $  \quad \text{NIQE} $\downarrow$  & \ FID$\downarrow$ \quad \text{NIQE} $\downarrow$  & \ FID$\downarrow$ \quad \text{NIQE} $\downarrow$     \\ \hline 
        PULSE & 64.86 \quad 5.10 & 102.74 \quad  5.23     & 86.45 \quad 5.14 & \quad - \qquad \ \  -   \\
        HiFaceGAN~\cite{Yang2020HiFaceGANFR} & 64.50 \quad 4.51 & 113.0 \quad \ \  4.86 & 116.1 \quad  4.88 & 79.15 \quad 5.06 \\ 
        PSFRGAN~\cite{ChenPSFRGAN} & 51.89 \quad 5.09 & 107.40 \quad  4.80   & 88.45 \quad 5.58 & 60.13 \quad 4.90 \\
        GPEN~\cite{2021gpen} & 54.85  \quad 3.90 & 106.90 \quad  4.08   & 80.70 \quad 4.34 & 58.78 \quad 4.30 \\ 
        GFPGAN~\cite{wang2021gfpgan} & 49.96 \quad 3.88 & 111.78 \quad 4.35    & 87.35 \quad 4.14 & 61.35 \quad 4.35    \\ 
        CodeFormer~\cite{zhou2022codeformer} & 52.02 \quad 4.01 & 116.23 \quad 4.98 & 83.19 \quad  4.70 &56.16 \quad 4.01  \\
        VQFR~\cite{gu2022vqfr}    & 50.64 \quad \textbf{3.59} & 105.18 \quad 3.94   &\textbf{75.38} \quad \textbf{3.61} & \textbf{54.01} \quad 3.99  \\ \hline 
        Ours-Simple & 49.81 \quad 3.95 & 111.18  \quad 4.08 & 77.89   \quad 3.75  & 57.13 \quad 4.14   \\
        Ours & \textbf{48.35} \quad 3.81 & \textbf{103.89} \quad \textbf{3.93} & 79.53 \quad 3.65 & 55.56 \quad \textbf{3.97}  \\ \hline
    \end{tabular}
    \vspace{-0.2cm}
    \caption{Quantitative comparison on the \textit{real-world} LFW-Test, CelebChild, WebPhoto and Casia-Test.}
    \vspace{-0.4cm}
    \label{tab:real_world_comparion}
\end{table*}

\begin{table}[h]
\centering
\label{tab:table4}
\resizebox{0.7\columnwidth}{!}{
    \begin{tabular}{ccccc}
    \hline
    $\mathcal{L}_{PER}$   &$\mathcal{L}_{ROI}$ & DA  & CAB  &FID$\downarrow$    \\
    \hline
    $\times$       & $\times$    & $\times$    & $\times$ & 27.94\\    % .  Done
    $\checkmark$   & $\times$   & $\times$   & $\times$ & 23.01 \\    %   Done
    $\checkmark$   & $\checkmark$      & $\times$   & $\times$ & 22.46 \\     %  
    $\checkmark$   &$\checkmark$       & $\checkmark$   & $\times$   & 22.86 \\     % 
    $\checkmark$   & $\checkmark$    & $\checkmark$  & $\checkmark$ & 20.51 \\    % 
    \hline
    \end{tabular}
}
\vspace{-0.25cm}
 \caption{FID between the different variants of BFRFormer} 
\label{tab:ablation_study}
\end{table}

\subsection{Ablation Study}
To better understand the roles of different components of BFRFormer and the training strategy, in this section, we conduct an ablation study by introducing some variants of BFRFormer and comparing with their BFR performance. 
\begin{table}[h]
\centering
\label{tab:tablea2}
\begin{tabular}{ccccc}
    \hline
    Method               & PSNR$\uparrow$  & FID$\downarrow$ &LPIPS $\downarrow$
    \\ \hline
    DFDNet~\cite{2020Blind}                  & 21.10    & 59.09  & 0.45 \\
    HiFaceGAN~\cite{Yang2020HiFaceGANFR}                 & 20.17    & 66.09    & 0.38       \\
    PSFRGAN~\cite{ChenPSFRGAN}\   & 19.86    & 62.05    & 0.32       \\
    GFPGAN~\cite{wang2021gfpgan} & 21.17   & 58.36   & 0.28     \\
    GPEN~\cite{2021gpen} & 19.85  & 59.70     & 0.29     \\ 
    VQFR~\cite{gu2022vqfr} & 20.51 & 58.01 &0.28  \\
    CodeFormer~\cite{zhou2022codeformer} & 22.18 & 60.62 &0.30 \\\hline
  %  RestoreFormer\cite{wang2022restoreformer} & 24.42 & 41.45 & 0.3650 \\\hline
    Ours  &\textbf{22.83}    &\textbf{57.37}          &\textbf{0.27 }   \\\hline
\end{tabular}
\vspace{-0.2cm}
\caption{Quantitative comparison (PNSR, FID, and LPIPS) of different BFR methods on Synthetic Datasets}
\vspace{-0.4cm}
\label{tab:compare_result}
\end{table}

\noindent \textbf{Baseline}.
The BFRFormer-simple uses a basic Transformer as the generator, without using the CAB module, perception loss, and local ROI loss, only using the basic loss functions: Adv loss, ID loss, and L1 loss. As shown in Table \ref{tab:ablation_study}, it can achieve high perceptual quality (low FID and NIQE) and produce surprisingly good results compared with GPEN~\cite{2021gpen}. Through attribution analysis, BFRFormer utilizes more information compared to GPEN.  \\
\textbf{Facial Component Discriminator}. Facial Component discriminators~\cite{wang2021gfpgan}, including right eyes, left eyes, and mouth, enhance the perceptually significant face components. Component discriminators with feature style loss could better capture the eye distribution and restore the plausible details~\cite{wang2021gfpgan}. The effectiveness is shown in Table  \ref{tab:ablation_study} and Figure \ref{fig:abality}.

\noindent \textbf{Data Augmentation}. Recent successes in Generative Adversarial Networks have affirmed the importance of using more data in GAN training. Augment data can improve the performance of the discriminator and generator. In this paper, the generated image and ground truth images are augmented by {Flipping, Color, Translation, Cutout} as in DiffAug. The effectiveness is shown in Table \ref{tab:ablation_study} and Figure \ref{fig:abality}.\\

% %%%%%%%%%%
\noindent \textbf{Channel Attention Block}. Local attention, nonetheless, sacrifices the ability to model long-range dependencies.
To activate more effective pixels for restoration tasks, channel attention utilizing more global information is used to further improve the performance of the Transformer. The effectiveness is shown in Table \ref{tab:ablation_study} and Figure \ref{fig:abality}.
\vspace{-10pt} 
\subsection{Comparison with State-of-the-art Methods}

\noindent \textbf{Synthetic Dataset}. We compare BFRFormer with several state-of-the-art face blind restoration methods, and the quantitative results are shown in Table \ref{tab:compare_result}. Our BFRFormer achieves better PNSR compared to other methods. Our results also obtain lower FID and LPIPS. BFRFormer can generate realistic and high-fidelity face images with details such as hair, eyes, mouth, etc. 

\noindent \textbf{Real World low-quality Datasets}. The final target of all methods is to restore authentic world LQ face images. To evaluate the generalization ability of different methods, we also compare the performance of BFRFormer on the LFW-Test, CelebChild-Test, WebPhoto-Test, and Casia-Test for evaluating the generalization of the proposed method. As shown in Table \ref{tab:real_world_comparion}, compared to previous face restoration methods on CelebAHQ-Test, BFRFormer gets better performance in FID and NIQE. 

\vspace{-0.3cm}
\section{Conclusion}
\vspace{-0.2cm}
In this paper, we propose a Transformer-based training method embedded in the GAN prior-based framework for blind face restoration. An aggregate attention module, composed of channel attention and double attention, is proposed to further improve generation quality. Extensive experiments on synthetic data and real-world low-quality images have demonstrated that BFRFormer can restore high-quality facial details while retaining the image background properly.

% -------------------------------------------------------------------------
\newpage

\bibliographystyle{IEEEbib}
\bibliography{strings}

\end{document}